\documentclass{amia}
\usepackage{lipsum}
\usepackage{booktabs}
\usepackage{tabularx}
\usepackage{multirow}
\usepackage{enumitem}
\usepackage{xcolor}
\usepackage{tikz}
\definecolor{softyellow}{RGB}{255, 236, 153} 
\usepackage{soul} 
\sethlcolor{softyellow}
\usepackage{array}
\usepackage{float} 
\usepackage{longtable}
\usepackage[most]{tcolorbox} 
\usepackage{caption}
\usepackage{hyperref}

\begin{document}

\title{A Dataset and Resources for Identifying Patient Health Literacy Information from Clinical Notes}

\author{Madeline Bittner$^1$, Dina  Demner-Fushman, MD, PhD$^1$, Yasmeen Shabazz$^2$, Davis Bartels$^1$, Dukyong Yoon, MD, PhD$^{2,3}$, Brad Quitadamo, BSN, RN, CCRN $^4$, Rajiv Menghrajani, MD$^5$, Leo Celi, MD$^{2,4}$, Sarvesh Soni, PhD$^1$, }

\institutes{
    $^1$National Library of Medicine, Bethesda, MD, USA; $^2$Massachusetts Institute of Technology (MIT), Cambridge, MA, USA; $^3$Yonsei University College of Medicine, Seoul, Republic of Korea;
    $^4$Beth Israel Deaconess Medical Center, Boston, Massachusetts, USA;
    $^5$NYC Health + Hospitals - Lincoln, Bronx, New York, USA
}

\maketitle

\section*{Abstract}

\textit{Health literacy is a critical determinant of patient outcomes, yet current screening tools are not always feasible and differ considerably in the number of items, question format, and dimensions of health literacy they capture, making documentation in structured electronic health records difficult to achieve. Automated detection from unstructured clinical notes offers a promising alternative, as these notes often contain richer, more contextual health literacy information, but progress has been limited by the lack of annotated resources. We introduce HEALIX, the first publicly available annotated health literacy dataset derived from real clinical notes, curated through a combination of social worker note sampling, keyword-based filtering, and LLM-based active learning. HEALIX contains 589 notes across 9 note types, annotated with three health literacy labels: low, normal, and high. To demonstrate its utility, we benchmarked zero-shot and few-shot prompting strategies across four open source large language models (LLMs). }

\section*{Introduction}
Health literacy is an individual's ability to obtain, understand, and use medical information to make informed decisions about their health\cite{iom04healthliteracy}. It strongly influences how well patients understand their diagnosis, manage their condition, and navigate the healthcare system\cite{oconnor20healthliteracyimpact}. According to The National Assessment of Adult Literacy Survey, 36\% of U.S. adults had basic or below average health literacy\cite{magnani18healthliteracystats}. Studies have shown that low health literacy is associated with more hospital visits, poorer health outcomes, and greater use of health care services\cite{shahid22healthoutcomes}. One study found that post operative patients with low health literacy were 62\% more likely to be readmitted than patients with baseline health literacy levels\cite{baker20surgery}. Patients with low health literacy are not able to successfully communicate with physicians or obtain information necessary to manage their condition\cite{schillinger04diabetes}. Consequently, health literacy substantially influences patient self-management and outcomes, making it a critical foundation of quality healthcare\cite{mccray05promotinghealthlit}.
However, a patient's health literacy level often remains unknown at the point of care.

Numerous validated health literacy screening tools are available to assess patients' health literacy; however, most require survey completion, which is neither feasible for acutely ill inpatients nor a priority for physicians managing their care\cite{cox17healthlitassociations}. Consequently, standardized health literacy measurements are seldom recorded in the electronic health records (EHRs) as structured data\cite{cristofori22healthliteracyreview}.
Instead, clinicians often document observations related to patients' health literacy in unstructured clinical narratives (e.g., patients' understanding of their condition and treatment plan or engagement in care). However, extracting this information through manual review is time-consuming and labor-intensive\cite{hu24jamianer}. Although clinical notes contain rich patient context, health literacy-related information within them remains largely overlooked and underutilized\cite{hoope25ptchar_fromEHRs}. This creates an opportunity for natural language processing (NLP) methods to automatically extract health literacy-related information from unstructured clinical notes, enabling more timely and scalable identification of patients' health literacy levels.

Automated extraction and classification of health literacy related statements from clinical notes has the potential to help physicians target educational interventions for patients who need them most. It can help develop better approaches to identifying patients with below baseline health literacy. Such identifications could enable interventions to reduce hospital readmission rates and improve patient outcomes. However, developing and validating such tools requires a robust, domain-specific dataset.

\begin{figure}[tbp]
    \centering
    \includegraphics[width=0.9\textwidth]{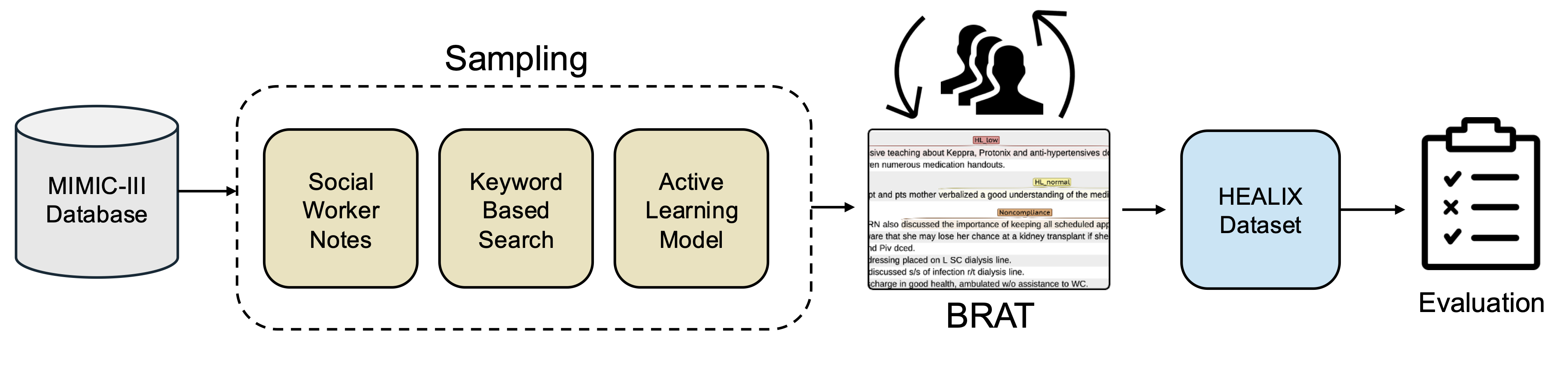}
    \caption{Workflow for dataset development. First, clinical notes were collected from the MIMIC-III database via three sampling strategies: random sampling of social worker notes, keyword-based filtering, and LLM-based active learning. Next, the sampled notes were independently annotated and reconciled to produce the HEALIX gold standard dataset. Finally, zero-shot and few-shot prompting strategies were applied to establish baseline model performance.}
    \label{fig:health_literacy_workflow}
\end{figure}

We address this gap by curating HEALIX, a dataset
of real clinical notes from a variety of domains, which we manually annotated with three fine-grained health literacy categories: low, normal, or high (Figure~\ref{fig:health_literacy_workflow}). To our knowledge, this is the first study to annotate specific health literacy indicators in clinical notes. To demonstrate the utility of HEALIX, we benchmarked the performance of multiple open-source models on a text classification task.
Instructions for accessing the dataset are available at \url{https://github.com/MaddieBitt/HEALIX}.

\section*{Background}
Health literacy is currently assessed in clinical settings using a variety of validated screening instruments; however, these tools lack a standardized questionnaire and protocol.\cite{liu18toolsforhealthliteracy} A review of eleven health literacy screening instruments found that each tool assessed different indicators of patient health literacy: some emphasized medical knowledge and terminology, others focused on communication or functional decision making skills, and questionnaires ranged from 16-64 questions\cite{liu18toolsforhealthliteracy}. Furthermore, no single questionnaire captures every aspect of health literacy, and some tools even incorporate numeracy evaluations, meaning current instruments do not account for all components of health literacy\cite{duell15optimal_hl_assessment}. This highlights the value of alternative approaches that draw on information already present in clinical documentation.

Despite growing recognition of health literacy's impact on patient outcomes, surveys indicate that health literacy is rarely documented in a structured format within EHRs. Instead, health literacy is often discussed by clinicians as indirect remarks scattered throughout narrative clinical notes or implied through statements about other clinical information\cite{cristofori22healthliteracyreview}. This mirrors the broader challenge of other social determinants of health (SDoH), which are similarly underrepresented in structured clinical data. LLMs have been applied to classify social determinants from clinical note sentences using multi-label sentence classification models\cite{keloth25SDoH}. We adopt a similar strategy for health literacy detection. 

Prior work in this area has focused on patient-generated text or structured EHR data. Schillinger et al. examines health literacy through patient secure messages to physicians, using linguistic features to predict patient health literacy levels\cite{schillinger21ECLIPPSE}. Similarly, Campbell et al. explored how other factors found in structured sections of EHRs can serve as a proxy for determining patient health literacy\cite{campbell19HL_structured}. However, neither of these approaches captures health literacy information as documented by unstructured clinical notes, which contains contextual observations and assessments of patient health literacy during patient visits. To our knowledge, no annotated datasets currently exist for extracting health literacy information from clinical notes. This work addresses these gaps by developing HEALIX, an annotated dataset of clinical notes labeled for patient health literacy, using a three-tier labeling system consistent with widely used validating screening instruments such as Newest Vital Sign (NVS) and Brief Health Literacy Screen (BHLS)\cite{weiss05_nvm_hlscreening, wallston14_briefhlscreen}, and demonstrates the utility of HEALIX through evaluation of multiple LLM classification tasks. 

\section*{Methods}

\subsection*{Data Collection}
We sourced clinical notes for HEALIX from the Medical Information Mart for Intensive Care (MIMIC III)\cite{johnson16mimic} corpus. We excluded all notes from Radiology, ECG, Echo, and Respiratory, as they rarely contained health literacy information. The initial set of clinical notes was seeded from a random sample of 300 social worker notes. Social worker notes were primarily selected because we anticipated that they would be more likely to contain relevant information about the patient's health literacy. To further expand HEALIX, we curated a list of broad keywords, partly based on our analysis of the social worker notes, to filter clinical notes most likely to contain health literacy instances.

The next sample was collected by taking a random sample of up to 3 notes for each keyword (Table~\ref{tab:keywords}). Clinical notes varied in length, ranging from an average of 18 to 142 sentences, and not every section in the notes contains health literacy information. Therefore, we extracted the sentence containing the keyword along with five preceding and five following sentences to capture the context relevant to health literacy while keeping the annotation workload manageable. Social worker notes, with an average of 18 sentences per note, were kept at their original length as they were short enough that sentence-level extraction was not necessary.  

An additional 200 notes were selected using an LLM based active learning approach to increase the sample diversity to include notes where the model exhibited greater uncertainty. We implemented a prompt-based active learning framework in which the LLaMA 3-8B-Instruct\cite{llama_8B_huggingface} model was prompted to read each clinical note and assign a health literacy label (Table~\ref{tab:llm_ALprompt}). Predictive entropy was then computed from the model's output probability distribution, and examples with the highest uncertainty were selected for annotation, following previous entropy-based active learning approaches\cite{xiang25promptAL}. We only prompted the model with 3 labels, ``poor", ``good" and ``not related", since our goal was to find additional notes for the underrepresented labels, while excluding the notes without health literacy information. The 200 notes with the highest entropy values were added to the existing pool for manual annotation. A sample of 15 notes was used for finalizing the annotation guidelines and was excluded from HEALIX, resulting in a final dataset of 589 notes.

\begin{table}[t]
\caption{Keywords used in health literacy dataset curation.}
\label{tab:keywords}
\centering
\footnotesize
\begin{tabular}{@{}
>{\raggedright\arraybackslash}p{0.31\textwidth}
>{\raggedright\arraybackslash}p{0.31\textwidth}
>{\raggedright\arraybackslash}p{0.31\textwidth}
@{}}
\toprule
adherence & insight (lack of/poor/little/minimal) & refusal \\
awareness & knowledge / limited knowledge & understands / comprehends \\
compliance / (non-)compliant & medical advice & understanding (good/poor/limited) \\
difficulty understanding & misunderstanding (of/the) & unaware (of diagnosis/history) \\
health literacy / literacy / literate & non(-)adherence / non(-)adherent & well(-)aware \\
inadequacy & patient choice / patient preference & \\
\bottomrule
\end{tabular}
\end{table}

\begin{table}[tbp]
\vspace{25pt}
\caption{LLaMA 3-8B Instruct prompt template for health literacy note level classification in the active learning workflow.}
\label{tab:llm_ALprompt}
\centering
\footnotesize
\begin{tcolorbox}[
  width=\textwidth,
  boxrule=0.4pt,
  colback=white,
  colframe=black,
  arc=2mm,
  left=3mm,right=3mm,top=2mm,bottom=2mm
]
\textbf{LLM Active Learning Prompt:}

\vspace{2mm}

You are a medical professional tasked with reading electronic medical records and assigning them one label from the list provided.

\vspace{2mm}

Choose exactly ONE label from:
\begin{itemize}[leftmargin=*, nosep]
  \item HL\_not\_related, HL\_poor, HL\_good
\end{itemize}

\vspace{2mm}

\textbf{Label definitions:}
\begin{itemize}[leftmargin=*, itemsep=2mm]
  \item \textbf{HL\_not\_related:} The document does not indicate the patient's health literacy level
  \item \textbf{HL\_poor:} The patient is unable to understand, interpret, and use health information to make informed medical decisions. The patient has difficulty understanding medical terminology or treatment plans, has a lack of insight into their condition, or lack of understanding of their current medical status. The patient refuses treatment, but is unaware of the expected outcome
  \item \textbf{HL\_good:} The patient has a good understanding of medical information and is able to make informed medical decisions. The patient is well aware of their medical status. The patient refuses the treatment plan, but is aware of the expected outcome.
\end{itemize}

\vspace{2mm}

\textbf{Additional rules:}
\begin{enumerate}[leftmargin=*, nosep]
  \item If the patient is physically or cognitively unable to make medical decisions, the label should correspond to the individual in the document who is making the medical decisions on behalf of the patient.
  \item Base your decision only on the text provided.
\end{enumerate}

\vspace{2mm}

\textbf{Output rules:}
Output EXACTLY ONE label (e.g., ``HL\_poor") and nothing else.

\end{tcolorbox}
\end{table}

\subsection*{Data Annotation}
Health literacy in unstructured data can come up as explicit terminology such as \textit{``poor health literacy''} or as more nuanced phrases or statements that would indicate a patient's health literacy such as \textit{``poor understanding of plan''}, \textit{``patient is aware of current condition''}, or \textit{``limited knowledge of diagnosis''}. Because of this nuanced language that can be used to describe health literacy, we developed strict guidelines for clinical note annotations. Annotators were asked to read the clinical note, assign the patient a health literacy level, and then label the sentences that contained information supporting that assigned level. Annotators were asked to label the sentences which directly or indirectly indicated the patient's health literacy level. If a patient is sedated, cognitively impaired, or has other reasons for not being able to make medical decisions for themselves, then the health literacy label goes to the individual who is in charge of the patient's medical care. To capture a range of health literacy levels, we annotated at three different levels, high, normal, or low, with notes containing no literacy indicator annotated as not related (Table~\ref{fig:definitions}). To distinguish between normal and high health literacy we defined high health literacy as the patient having demonstrated their understanding in a way that goes beyond a verbal confirmation of understanding of diagnosis or treatment plan. Each clinical note was independently annotated by two annotators, and disagreements at both the sentence and note levels were reconciled to create the HEALIX gold standard dataset.

\begin{table}[tbp]
\centering
\renewcommand{\arraystretch}{1.2}
\caption{Definitions of health literacy levels and examples.}
\label{fig:definitions}
\begin{tabularx}{\textwidth}{@{} l p{0.54\textwidth} p{0.34\textwidth} @{}}
\hline
\textbf{Level} & \textbf{Definition} & \textbf{Examples} \\
\hline
Low & Patient is unable to understand, interpret, and use health information to make decisions & \textit{The patient does not understand the need to follow up with physician.}
\\ \hline
Normal & An individual is able to obtain, understand, and use health/medical information to make informed health related decisions & \textit{The patient is aware of the treatment plan.}
\\
\hline
High & Patient demonstrates an above average understanding of medical related information and is able to use that information to make informed decisions & \textit{The patient demonstrated their medication regimen.} 
\\
\hline
\end{tabularx}
\end{table}

\subsection*{Models}
To evaluate HEALIX, we developed zero-shot and few-shot prompting strategies for a multiclass text classification task using the three health literacy labels. As a baseline, we used an SVM classifier shown comparable to neural approaches for some clinical text classification tasks\cite{wang2019clinical}. We used Qwen and LLaMa models (Qwen3-8B\cite{qwen3_8b_huggingface}, Qwen3-32B\cite{qwen3_32b_huggingface}, LLaMa3-8B-Instruct\cite{llama_8B_huggingface}, and LLaMa3.3-70B\cite{llama70B_huggingface}) for our evaluation, as these models have been successfully implemented in similar studies\cite{qian26extractinglanguageinfo}. All models were evaluated in their pretrained state without any task specific fine tuning. Each model was provided with a structured prompt, instructing it to read the entire clinical note and, using the full note context, assign a health literacy label to the patient (Table~\ref{tab:modelprompts}). We randomly allocated 20\% of HEALIX for prompt development and optimization, and reserved the remaining 80\% for final evaluation, with both splits preserving the label distribution. We evaluated the models using two labeling schemes. Strict evaluation used the three distinct health literacy categories defined in Table~\ref{fig:definitions}. Lenient evaluation collapsed the \textit{high} and \textit{normal} labels into a single \textit{``good''} category, retained \textit{low} as \textit{``poor''}, and kept \textit{not related} as a separate category.
For lenient evaluation, we applied this mapping to both the gold labels and model predictions before computing metrics.

\begin{table}[tbp] 
\caption{Prompt template used for zero-shot and few-shot clinical note strict classification during model evaluation.}
\centering
\label{tab:modelprompts}
\begin{tcolorbox}[
  width=\textwidth,
  boxrule=0.4pt,
  colback=white,
  colframe=black,
  arc=2mm,
  left=2mm,right=2mm,top=2mm,bottom=2mm,
  fontupper=\footnotesize  
]
\textbf{Health Literacy Classification Prompt:}

\vspace{1mm}

Health literacy is defined as the level to which an individual is able to obtain, process, and understand basic health information needed to make appropriate health decisions. You are a medical professional and your task is to classify each of the following clinical notes based on the patient's health literacy level.

\vspace{1mm}

\textbf{Important Classification Rule:}

The health literacy label should be based on explicit evidence found in the clinical note. The evidence to label the note may be one sentence within the clinical note or more than one sentence within the clinical note. Carefully examine the sentences surrounding the evidence for context before selecting a label.

\vspace{1mm}

\textbf{Labels:} HL\_not\_related, HL\_low, HL\_normal, HL\_high

\vspace{1mm}

\textbf{Labels and Definitions:}

\vspace{1mm}

\textbf{HL\_high:}

\textit{Definition:} The patient IS ABLE to comprehend medical information to make informed decisions about their personal health. The patient's GOOD understanding of the diagnosis and treatment plan is emphasized in the clinical note. The patient demonstrates informed decision making capacity, weighing treatment risks and benefits, and selecting treatment options that are most aligned with their personal goals. This is an example of evidence that would indicate HIGH HEALTH LITERACY. (e.g., The physician explained the medication list and use to the patient. The patient demonstrates good understanding of medication regimen. The patient was discharged home with wife.)\\
\vspace{0.5mm}
\textbf{[Inserted High Health Literacy Clinical Note]}

\vspace{1mm}

\textbf{HL\_normal:}

\textit{Definition:} The patient IS ABLE to verbalize a GOOD understanding of the diagnosis and/or treatment plan. Given the context of surrounding sentences, the evidence IS related to the patient's ability to obtain and use medical information to make informed medical decisions. The patient's GOOD understanding of the diagnosis and treatment plan is stated matter-of-factly in the note, but NOT emphasized. The actions described in the note SHOW that the patient IS aware of their condition. The patient IS ABLE to make informed medical decisions about their treatment plan. This is an example of evidence that would indicate NORMAL HEALTH LITERACY. (e.g., In the outpatient unit the physician checks the patient's wounds and prepares the patient for discharge. The patient verbalizes understanding to continue with prescribed medications and follow up appointments following surgery. The patient was diagnosed with the husband and children.)\\
\vspace{0.5mm}
\textbf{[Inserted Normal Health Literacy Clinical Note Section]}

\vspace{1mm}

\textbf{HL\_low:}

\textit{Definition:} The patient DOES NOT understand medical information and is NOT able to make informed decisions about their personal health. The patient DOES NOT understand their treatment plan or current medical state. This is an example of evidence that would indicate LOW HEALTH LITERACY. (e.g., During the home visit the social worker indicated that the patient's condition had worsened. SW stated that the patient doesn't understand the importance of taking his medication regularly. The patient will likely need readmission.)

\textit{Note:} If the evidence for HL\_low is related to a mental condition or the patient's cognitive function, this is considered HL\_not\_related. This is considered a cognitive disorder and is not related to the patient's ability to understand medical concepts.\\
\vspace{0.5mm}
\textbf{[Inserted Low Health Literacy Clinical Note Section]}

\vspace{1mm}

\textbf{HL\_not\_related:}

\textit{Definition:} The note does NOT contain any information about the patient's ability to make informed decisions about their personal health. The note might contain evidence of health literacy, BUT given the surrounding sentences, is NOT related to health literacy. Any evidence that is a boilerplate response from the medical professional (e.g., Pt understands need for surgery, or Patient was provided education and understands) is NOT considered related to health literacy. Evidence within the note that is related to a patient's literacy level, their ability to read and write, is also considered not related to health literacy. This is an example of evidence that would NOT indicate a patient's health literacy level. (e.g., Patient presented to the hospital with SOB. The patient has a history of left lobe pleural effusions. The patient understands the test and imaging needed for diagnosis.)

\textit{Note:} Evidence, and surrounding context, that describes the patient's mental state is considered HL\_not\_related. This is not related to their ability to understand medical concepts.\\
\vspace{0.5mm}
\textbf{[Inserted Not Health Literacy Related Clinical Note Section]}

\vspace{1mm}

\textbf{When uncertain between HL\_normal and HL\_high:}

If the text shows ANY of the following, classify the clinical note as HL\_high:
\begin{itemize}[leftmargin=*, nosep]
  \item The patient uses critical thinking when making a medical decision
  \item The patient is demonstrating their understanding of the condition or treatment plan through actions or through substantive discussion, informed questions, or decision-making that indicates advanced grasp of their medical situation
\end{itemize}

\vspace{1mm}

\textbf{Additional Rules:} If the patient is physically or cognitively unable to make medical decisions, the label should correspond to the individual in the document who is making the medical decisions on behalf of the patient.
\vspace{1mm}

\textbf{Output rules:} Your entire response must be exactly one of: HL\_not\_related, HL\_low, HL\_normal, HL\_high, Do NOT include any other words, Do NOT include explanations, Do NOT write sentences

\vspace{1mm}

\end{tcolorbox}
\end{table}

\section*{Results}

\subsection*{Inter Annotator Agreement}
We computed unweighted F scores as agreement statistics between annotators. Inter-annotator agreement was calculated under both strict and lenient label mappings. The note-level strict agreement F-score was 0.52, which increased to 0.63 under the lenient agreement scheme. The increase in agreement between strict and lenient measurements reflects the inherent subjectivity in this annotation task, as higher agreement when collapsing \textit{high} and \textit{normal} labels suggests that distinguishing between the two is a highly context-dependent task. At a more granular level, we examined sentence-level agreement for sentences that both annotators labeled for health literacy. There were 168 sentences that both annotators labeled with a health literacy category (\textit{Low}, \textit{Normal}, \textit{High}). At this level, the strict agreement F-Score was 0.72 and the lenient agreement F-Score was 0.93. This indicates that when both annotators identified a sentence as containing health literacy evidence, they had high agreement on the health literacy level conveyed. 


\begin{table}[tbp]
\centering
\renewcommand{\arraystretch}{1.1}
\caption{Model Performance Evaluation at the strict, 4 classes level, and lenient, 3 classes level.}
\label{tab:model_evaluations}
\begin{tabularx}{\textwidth}{@{} l l *{6}{>{\centering\arraybackslash}X} @{}}
\hline
\multirow{2}{*}{\textbf{Model}} & \multirow{2}{*}{\textbf{Setting}} & \multicolumn{3}{c}{\textbf{Strict}} & \multicolumn{3}{c}{\textbf{Lenient}} \\
\cline{3-5} \cline{6-8}
 & & \textbf{P} & \textbf{R} & \textbf{F1} & \textbf{P} & \textbf{R} & \textbf{F1} \\
\hline
SVM & -- & 0.64 & 0.39 & 0.41 & 0.56 & 0.49 & 0.50 \\
\hline
\multirow{2}{*}{LLaMA 3-8B-Instruct}    & Zero-Shot & 0.46 & 0.47 & 0.46 & 0.55 & 0.54 & 0.54 \\
                                         & Few-Shot  & 0.45 & 0.48 & 0.45 & 0.55 & 0.55 & 0.55 \\
\hline
\multirow{2}{*}{LLaMA 3.3-70B-Instruct} & Zero-Shot & 0.52 & 0.51 & \textbf{0.51} & 0.64 & 0.62 & \textbf{0.63} \\
                                         & Few-Shot  & 0.47 & 0.49 & 0.46 & 0.61 & 0.62 & 0.60 \\
\hline
\multirow{2}{*}{Qwen 3-8B}              & Zero-Shot & 0.58 & 0.49 & 0.49 & 0.70 & 0.57 & 0.57 \\
                                         & Few-Shot  & 0.53 & 0.49 & 0.47 & 0.66 & 0.57 & 0.58 \\
\hline
\multirow{2}{*}{Qwen-3-32B}             & Zero-Shot & 0.49 & 0.50 & 0.47 & 0.63 & 0.58 & 0.59 \\
                                         & Few-Shot  & 0.47 & 0.50 & 0.46 & 0.60 & 0.59 & 0.59 \\
\bottomrule
\end{tabularx}
\end{table}

\begin{figure}[tbp]
    \centering
    \includegraphics[width=\linewidth]{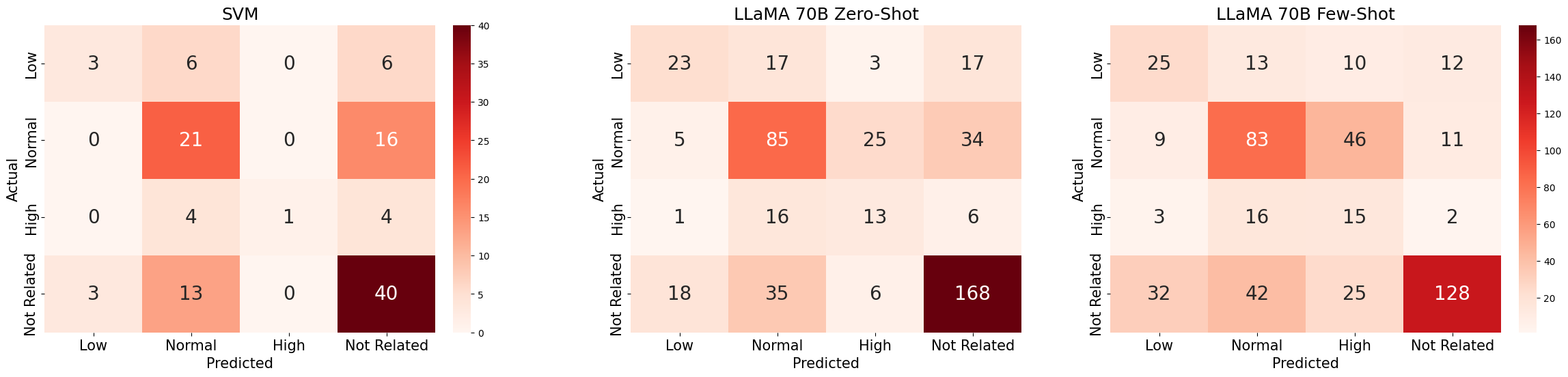}
    \caption{Confusion matrices for SVM and LLaMA 70B (zero-/few-shot). SVM matrix uses a different color bar than the LLaMA matrices.}
    \label{fig:confusion_matrices}
\end{figure}

\subsection*{Dataset Characteristics}
Table~\ref{tab:patient_demographics} summarizes the distribution of patient demographics, encounter characteristics, and note categories across
HEALIX's 589 clinical notes spanning nine note types, including Social Work, Nursing, Discharge, Physician, and Rehab Services. HEALIX consists of 45 high, 186 normal, 75 low health literacy notes, with 285 not related notes. HEALIX is primarily composed of emergency room patients (80.14\%), who are predominantly white (68.59\%) and English-speaking (67.06\%), with Spanish being the second most common language (3.06\%). The largest age group represented is 60-69 years (20.71\%), and the majority of patients carry either private (40.41\%) or Medicare (39.56\%) insurance. Note that some patients may appear in the demographics more than once. 

\renewcommand{\arraystretch}{1.2}
\begin{table}[tbp]
\centering
\caption{Patient demographics, encounter characteristics, and clinical note categories in the dataset, stratified by health literacy label (Low, Normal, High, Not Related) and overall (Total). Values are shown as count (\%) within column, with percentages rounded to one decimal place.}
\label{tab:patient_demographics}
\resizebox{\linewidth}{!}{
\begin{tabular}{@{}p{2cm} p{3.8cm} r r r r r@{}}
\hline
{\footnotesize\textbf{Category}} & {\footnotesize\textbf{Value}} & {\footnotesize\textbf{Low (N=75)}} & {\footnotesize\textbf{Normal (N=186)}} & {\footnotesize\textbf{High (N=45)}} & {\footnotesize\textbf{Not Related (N=283)}} & {\footnotesize\textbf{Total (N=589)}} \\
\hline

\multirow{2}{*}{Sex} 
  & Male   & 49 (65.3) & 112 (60.2) & 29 (64.4) & 143 (50.5) & 333 (56.5) \\ \cline{2-7}
  & Female & 26 (34.7) & 74 (39.8) & 16 (35.6) & 140 (49.5) & 256 (43.5)\\\midrule[\heavyrulewidth]

\multirow{5}{*}{Ethnicity}
  & White & 47 (62.7) & 124 (66.7) & 37 (82.2) & 196 (69.3) & 404 (68.6) \\ \cline{2-7}
  & Black/African American & 16 (21.3) & 17 (9.1) & 1 (2.2) & 28 (9.9) & 62 (10.5) \\ \cline{2-7}
  & Hispanic/Latino & 3 (4.0) & 16 (8.6) & 3 (6.7) & 17 (6.0) & 39 (6.6) \\ \cline{2-7}
  & Asian & 1 (1.3) & 8 (4.3) & 3 (6.7) & 11 (3.9) & 23 (3.9) \\ \cline{2-7}
  & Other/Decline to Answer & 8 (10.7) & 21 (11.3) & 1 (2.2) & 31 (11.0) & 61 (10.4) \\
  \midrule[\heavyrulewidth]

\multirow{3}{*}{Language}
  & English & 58 (77.3) & 99 (53.2) & 23 (51.1) & 215 (76.0) & 395 (67.1) \\ \cline{2-7}
  & Spanish & 2 (2.7) & 7 (3.8) & 0 (0.0) & 9 (3.2) & 18 (3.1) \\ \cline{2-7}
  & Other/Decline to Answer & 15 (20.0) & 80 (43.0) & 22 (48.9) & 59 (20.9) & 176 (29.9) \\ \midrule[\heavyrulewidth]

\multirow{5}{*}{Insurance}
  & Private & 17 (22.7) & 88 (47.3) & 28 (62.2) & 105 (37.1) & 238 (40.4) \\ \cline{2-7}
  & Medicare & 36 (48.0) & 68 (36.6) & 12 (26.7) & 117 (41.3) & 233 (39.6) \\ \cline{2-7}
  & Medicaid & 18 (24.0) & 19 (10.2) & 4 (8.9) & 44 (15.6) & 85 (14.4) \\ \cline{2-7}
  & Government & 4 (5.3) & 9 (4.8) & 1 (2.2) & 9 (3.2) & 23 (3.9) \\ \cline{2-7}
  & Self Pay & 0 (0.0) & 2 (1.1) & 0 (0.0) & 8 (2.8) & 10 (1.7) \\ \midrule[\heavyrulewidth]

\multirow{8}{*}{Age}
  & 0-17  & 3 (4.0)    & 50 (26.9) & 12 (26.7) & 15 (5.3)   & 80 (13.6)  \\ \cline{2-7}
  & 18-29 & 5 (6.7)    & 7 (3.8)   & 2 (4.4)   & 22 (7.8)   & 36 (6.1)   \\ \cline{2-7}
  & 30-39 & 9 (12.0)   & 9 (4.8)   & 0 (0.0)   & 21 (7.4)   & 39 (6.6)   \\ \cline{2-7}
  & 40-49 & 9 (12.0)   & 26 (14.0) & 6 (13.3)  & 40 (14.1)  & 81 (13.8)  \\ \cline{2-7}
  & 50-59 & 11 (14.7)  & 23 (12.4) & 5 (11.1)  & 47 (16.6)  & 86 (14.6)  \\ \cline{2-7}
  & 60-69 & 21 (28.0)  & 31 (16.7) & 10 (22.2) & 60 (21.2)  & 122 (20.7) \\ \cline{2-7}
  & 70-79 & 9 (12.0)   & 24 (12.9) & 4 (8.9)   & 34 (12.0)  & 71 (12.1)  \\ \cline{2-7}
  & 80+   & 8 (10.7)   & 16 (8.6)  & 6 (13.3)  & 44 (15.6)  & 74 (12.6)  \\
  \midrule[\heavyrulewidth]

\multirow{4}{*}{\shortstack[l]{Admission\\Type}}
  & Emergency & 67 (89.3) & 130 (69.9) & 28 (62.2) & 247 (87.3) & 472 (80.1) \\ \cline{2-7}
  & Newborn & 3 (4.0) & 48 (25.8) & 12 (26.7) & 12 (4.2) & 75 (12.7) \\ \cline{2-7}
  & Elective & 3 (4.0) & 7 (3.8) & 5 (11.1) & 18 (6.4) & 33 (5.6) \\ \cline{2-7}
  & Urgent & 2 (2.7) & 1 (0.5) & 0 (0.0) & 6 (2.1) & 9 (1.5) \\ \midrule[\heavyrulewidth]

\multirow{4}{*}{\shortstack[l]{Admission\\Location}}
  & Emergency Room & 50 (66.7) & 92 (49.5) & 18 (40.0) & 168 (59.4) & 328 (55.7) \\ \cline{2-7}
  & Referral & 13 (17.3) & 65 (35.0) & 20 (44.4) & 64 (22.6) & 162 (27.5) \\ \cline{2-7}
  & Transfer & 12 (16.0) & 28 (15.1) & 7 (15.6) & 51 (18.0) & 98 (16.6) \\ \cline{2-7}
  & Not Available & 0 (0.0) & 1 (0.5) & 0 (0.0) & 0 (0.0) & 1 (0.2) \\
  \midrule[\heavyrulewidth]

\multirow{6}{*}{\shortstack[l]{Note\\Category}}
  & Social Work & 40 (53.3) & 81 (43.6) & 15 (33.3) & 161 (56.9) & 297 (50.4) \\ \cline{2-7}
  & Nursing & 10 (13.3) & 84 (45.2) & 20 (44.4) & 45 (15.9) & 159 (27.0) \\ \cline{2-7}
  & Discharge Summary & 20 (26.7) & 17 (9.1) & 5 (11.1) & 56 (19.8) & 98 (16.6) \\ \cline{2-7}
  & Physician & 3 (4.0) & 2 (1.1) & 0 (0.0) & 13 (4.6) & 18 (3.1) \\ \cline{2-7}
  & Rehab Services & 2 (2.7) & 1 (0.5) & 5 (11.1) & 5 (1.8) & 13 (2.2) \\ \cline{2-7}
  & Other & 0 (0.0) & 1 (0.5) & 0 (0.0) & 3 (1.1) & 4 (0.7) \\
  \midrule[\heavyrulewidth]
\end{tabular}}
\end{table}

\section*{Model Evaluations}
Macro average evaluation metrics were used to measure precision, recall, and F1, and the results are shown in Table~\ref{tab:model_evaluations}. The SVM model yielded the lowest F1 scores across both strict and lenient evaluations, underperforming all LLM base models. The LLaMA 3.3-70B Instruct model performed the best overall, with the highest F1 scores in a majority of cases. The highest F1 score was 0.63, from the LLaMA 3.3-70B Instruct model in the lenient evaluation. Looking at the model prediction results shown in Figure~\ref{fig:confusion_matrices}, the zero-shot model struggled most with distinguishing between normal and not related labels. The few-shot model showed similar confusion between normal and not related, as well as low and not related, with misclassifications most frequent between high and normal labels. This pattern of high and normal confusion motivated the lenient evaluation that collapsed these two categories into a single label.

\section*{Discussion}
We curated HEALIX, a health literacy dataset using clinical notes within the MIMIC database. To our knowledge, this is the first publicly available annotated health literacy dataset derived from clinical narrative notes. HEALIX contains 589 clinical notes from 564 individual patients across 9 note categories, serving as a foundational resource for training and evaluating NLP models for health literacy detection in clinical notes. 

Most low health literacy instances in HEALIX were identified within social worker and discharge summary notes, which contain more unstructured narrative sections that tend to capture patient behavior, context, and knowledge rather than clinical observations alone. Additionally, health literacy indicators in these sections were more likely to appear as indirect, narrative, or action-based cues rather than explicit statements (e.g \textit{``history of avoidance of medical care''}, \textit{``coping appropriately with aftermath''}, \textit{``some insight into limitations''}). Other studies have similarly found that medical professionals are more likely to document health literacy-relevant information in the unstructured narrative test, which suggests that these notes carry significant diagnostic value that structured fields fail to capture\cite{cristofori22healthliteracyreview}. Given this, unstructured clinical notes represent a critical yet often underutilized resource for inferring patient health literacy. 

Patient demographics in HEALIX indicate that individuals aged 60-69  and Black/African American populations exhibit lower health literacy levels, which is consistent with existing literature demonstrating that minority populations\cite{lima24hl_factors} and older adults\cite{chesser16healthlit_age} are disproportionately affected by low health literacy. While our findings agree with the literature, focused studies are needed to determine if the health literacy differences are genuine or uncover a documentation bias (clinicians noting literacy concerns more for certain populations).

Our primary focus was to develop HEALIX and establish benchmark model evaluation metrics, and these results should be interpreted as baseline. The zero-shot and few-shot models showed promise, and prior work suggests that fine tuning pretrained models may further improve performance\cite{nunes24finetuning}. The models performed better when health literacy was conveyed through direct statements about patient understanding, but struggled with more nuanced, indirect examples where broader note context was necessary to make an accurate classification. As shown in Figure~\ref{fig:annotation_example}, in Example 1, the patient's high health literacy is conveyed directly (i.e., the note states that the patient understands the procedure), and both the annotators and model labeled that note as high health literacy. In Example 2, normal health literacy is conveyed indirectly through the families inquiry into alternative treatment options, yet the model labeled the note as not health literacy related. In Example 3, although the family demonstrated partial understanding of the patient's condition, the broader context of the note reveals a substantial gap in understanding the quality-of-life implications, which is a nuance the model failed to capture. These findings indicate that health literacy detection in clinical notes requires a contextual, note-level understanding, rather than explicit statements alone. 

\begin{figure}[tbp]
\centering
\begin{tcolorbox}[width=0.8\textwidth, colback=white, colframe=black, boxrule=0.5pt]
\raggedright

\textbf{Clinical Note Example 1:}

share written procedure on ASA desensitization. \hl{Understands pt. role in producure. Both wife and pt. state they remember the heart cath here and have no questions regarding same}. Briefly reviewed interventions of stent prior to cath lab.

\vspace{0.5em}
\vspace{0.5em}

\textbf{Annotators: HL\_high} 
\hfill
\textbf{Model: HL\_high}

\vspace{0.8em}
\hrule
\vspace{0.8em}

\textbf{Clinical Note Example 2:}

\hl{Son inquiring if pt could be candidate for kidney transplant and could he be donor.} MDs informed son that this was possibility but in the future. 2) Concern for pt pain control they feel pt being over medicated and is sensitive to pain meds. Palliative care MD informed family that they will try fentanyl patch, to which family agreed. 3) Plan going forward for pt. \hl{Family aware of reported need for amputation surgery}

\vspace{0.5em}
\vspace{0.5em}

\textbf{Annotators: HL\_normal} 
\hfill
\textbf{Model: HL\_not\_related}

\vspace{0.8em}
\hrule
\vspace{0.8em}

\textbf{Clinical Note Example 3:}

Family repeated back their understanding of pt s condition and likely future course of treatment based on information given to them at the family mtg that occurred on Wednesday. They demonstrated a fairly accurate understanding of pt s condition; however, \hl{their reporting reflected lack of understanding of the potential quality of life issues pt could face} if she ends up going to a rehab facility w/a long-term trach/peg.

\vspace{0.5em}
\vspace{0.5em}

\textbf{Annotators: HL\_low} 
\hfill
\textbf{Model: HL\_normal}

\end{tcolorbox}
\caption{Annotated clinical note excerpts illustrating examples of varying model classification difficulty. Highlighted text represents evidence of patient health literacy within the note. Any misspellings or abbreviations are reported as they appear in the original note.}
\label{fig:annotation_example}
\end{figure}

This study has several limitations. HEALIX is derived from a single source (MIMIC), which contains patients from one hospital and may not be representative of broader patient populations. Additionally, the demographic composition of HEALIX may limit generalizability of our findings related to health literacy disparities across different populations. Some of the clinical notes in HEALIX consist of extracted excerpts rather than full notes, to capture health literacy evidence and its surrounding context while filtering out noise. However, this approach may inadvertently omit broader context from the remainder of the note that could be relevant for health literacy classification. Notably, the models evaluated in this study were not fine tuned, and future work exploring fine tuning may result in stronger performance. 

\subsection*{Conclusion}
Health literacy plays a significant role in determining patient outcomes, yet formal screenings are not always feasible. To support automated detection, we developed HEALIX, the first publicly available annotated health literacy dataset derived from clinical notes. HEALIX contains 589 notes across 9 note types annotated with three health literacy labels: low, normal, and high. Benchmark evaluation of zero-shot and few-shot models showed promise but highlighted challenges with context-dependent instances. HEALIX serves as a foundational resource for training and evaluating NLP models for health literacy detection in clinical notes, with the broader goal of improving patient health outcomes. 

\subsection*{Acknowledgments}
This research was supported by the Intramural Research Program of the National Institutes of Health (NIH) and utilized the computational resources of the NIH HPC Biowulf cluster (http://hpc.nih.gov). The contributions of the NIH authors are considered Works of the United States Government. The findings and conclusions presented in this paper are those of the authors and do not necessarily reflect the views of the NIH or the U.S. Department of Health and Human Services.

\makeatletter
\renewcommand{\@biblabel}[1]{\hfill #1.}
\makeatother

\bibliographystyle{vancouver}
\bibliography{amia}  

@article{wang2019clinical,
  title={A clinical text classification paradigm using weak supervision and deep representation},
  author={Wang, Yanshan and Sohn, Sunghwan and Liu, Sijia and Shen, Feichen and Wang, Liwei and Atkinson, Elizabeth J and Amin, Shreyasee and Liu, Hongfang},
  journal={BMC medical informatics and decision making},
  volume={19},
  number={1},
  pages={1},
  year={2019},
  publisher={Springer}
}

@article{johnson16mimic,
author                = {{Johnson, Alistair E. W. and Pollard, Tom J. and Shen, Lu and Lehman, Li-wei H. and Feng, Mengling and Ghassemi, Mohammad and Moody, Benjamin and Szolovits, Peter and Celi, Leo Anthony and Mark, Roger G.}},
title                 = {MIMIC-III, a freely accessible critical care database},
journal               = {Scientific Data},
volume                = {3},
pages                 = {160035},
year                  = {2016}
}

@article{schillinger04diabetes,
author                = {{Dean Schillinger and Andrew Bindman and Frances Wang and Anita Stewart and John Piette}},
title                 = {Functional health literacy and the quality of physician–patient communication among diabetes patients},
journal               = {Patient Education and Counseling},
volume                = {52},
pages                 = {315-323},
year                  = {2004}
}

@book{iom04healthliteracy,
  author    = {{Institute of Medicine (US) Committee on Health Literacy}},
  title     = {{Health Literacy: A Prescription to End Confusion}},
  address   = {Washington (DC)},
  publisher = {National Academies Press (US)},
  year      = {2004},
  pmid      = {25009856}
}

@article{shahid22healthoutcomes,
  author  = {Shahid, R. and Shoker, M. and Chu, L. M. and Frehlick, R. and Ward, H. and Pahwa, P.},
  title   = {{Impact of low health literacy on patients' health outcomes: a multicenter cohort study}},
  journal = {BMC Health Services Research},
  volume  = {22},
  pages   = {1148},
  year    = {2022}
}

@article{hu24jamianer,
    author = {Hu, Yan and Chen, Qingyu and Du, Jingcheng and Peng, Xueqing and Keloth, Vipina Kuttichi and Zuo, Xu and Zhou, Yujia and Li, Zehan and Jiang, Xiaoqian and Lu, Zhiyong and Roberts, Kirk and Xu, Hua},
    title = {Improving large language models for clinical named entity recognition via prompt engineering},
    journal = {JAMIA},
    volume = {31},
    number = {9},
    pages = {1812-1820},
    year = {2024}
}

@article{magnani18healthliteracystats,
  author  = {{Magnani JW and Mujahid MS and Aronow HD and Cené CW and Dickson VV and Havranek E and Morgenstern LB and Paasche-Orlow MK and Pollak A and Willey JZ}},
  journal = {Circulation},
  pages   = {e48--e74},
  title   = {{Health Literacy and Cardiovascular Disease: Fundamental Relevance to Primary and Secondary Prevention: A Scientific Statement From the American Heart Association}},
  volume  = {138},
  year    = {2018},
}

@article{oconnor20healthliteracyimpact,
  author  = {{O'Conor R and Moore A and Wolf MS}},
  journal = {Stud Health Technol Inform},
  pages   = {3--21},
  title   = {{Health Literacy and Its Impact on Health and Healthcare Outcomes}},
  volume  = {269},
  year    = {2020},
}

@article{baker20surgery,
  author = {Samantha Baker and Emily Malone and Laura Graham and Elise Dasinger and Tyler Wahl and Ashley Titan and Joshua Richman and Laurel Copeland and Edith Burns and Jeffrey Whittle and Mary Hawn and Melanie Morris},
  journal = {The American Journal of Surgery},
  pages = {1138-1144},
  title = {Patient-reported health literacy scores are associated with readmissions following surgery},
  volume = {220},
  year = {2020}
}

@article{qian26extractinglanguageinfo,
  author = {Lingfei Qian and Na Hong and Yujia Zhou and Qianqian Xie and Ruey-Ling Weng and Pitchaya Chairuengjitjaras and Xinsong Du and John Lian and Gad A. Marshall and Suzanne V. Blackley and John Novoa-Laurentiev and Yakeel T. Quiroz and Tae Youn Kim and Nicole Adams and Michelle L. Dossett and Li Zhou and Hua Xu},
  journal = {International Journal of Medical Informatics},
  pages = {106116},
  title = {Extracting language information from clinical notes using large language models},
  volume = {205},
  year = {2026},
}

@article{liu18toolsforhealthliteracy,
  author  = {{Liu H and Zeng H and Shen Y and Zhang F and Sharma M and Lai W and Zhao Y and Tao G and Yuan J and Zhao Y}},
  journal = {Int. J. Environ. Res. Public Health},
  title   = {{Assessment Tools for Health Literacy among the General Population: A Systematic Review}},
  pages   = {1711},
  volume  = {15},
  year    = {2018},
}

@article{cox17healthlitassociations,
    author = {Sarah R. Cox and Michael G. Liebl and Meghan N. McComb and Jason Q. Chau and Allison A. Wilson and May Achi and Kevin W. Garey and David Wallace},
    journal = {Research in Social and Administrative Pharmacy},
    title = {Association between health literacy and 30-day healthcare use after hospital discharge in the heart failure population},
    pages = {754-758},
    volume = {13},
    year = {2017}
}

@article{schillinger21ECLIPPSE,
  author    = {Schillinger, Dean and Balyan, Ramin and Crossley, Scott A. and McNamara, Danielle S. and Liu, Jing Y. and Karter, Andrew J.},
  journal   = {Health Services Research},
  title     = {Employing computational linguistics techniques to identify limited patient health literacy: Findings from the ECLIPPSE study},
  pages     = {132--144},
  volume    = {56},
  year      = {2021},
}

@article{keloth25SDoH,
  author  = {Keloth, Vipina K. and Selek, Salih and Chen, Qingyu and others},
  journal = {npj Digital Medicine},
  title   = {Social determinants of health extraction from clinical notes across institutions using large language models},
  pages   = {287},
  volume  = {8},
  year    = {2025},
}

@article{cristofori22healthliteracyreview,
  author  = {Cristofori, E. and Zeffiro, V. and Alvaro, R. and D'Agostino, F. and Zega, M. and Cocchieri, A.},
  journal = {SAGE Open Nursing},
  title   = {Health Literacy in Patients’ Clinical Records of Hospital Settings: A Systematic Review},
  year    = {2022},
  volume  = {8},
  pages   = {23779608221078555},
  doi     = {10.1177/23779608221078555}
}

@article{nunes24finetuning,
  author       = {Nunes, Miguel and Bone, Joao and Ferreira, Joao C and Elvas, Luis B},
  title        = {Health Care Language Models and Their Fine-Tuning for Information Extraction: Scoping Review},
  journal      = {JMIR Medical Informatics},
  year         = {2024},
  volume       = {12},
  pages        = {e60164},
}

@article{lima24hl_factors,
title = {Factors associated with poor health literacy in older adults: A systematic review},
author = {Ana Caroline Pinto Lima and Madson Alan Maximiano-Barreto and Tatiana Carvalho Reis Martins and Bruna Moretti Luchesi},
journal = {Geriatric Nursing},
volume = {55},
pages = {242-254},
year = {2024}
}

@article{xiang25promptAL,
title = {PromptAL: Sample-aware dynamic soft prompts for few-shot active learning},
journal = {Knowledge-Based Systems},
volume = {329},
pages = {114354},
year = {2025},
author = {Hui Xiang and Jinqiao Shi and Ting Zhang and Xiaojie Zhao and Yong Liu and Yong Ma},

}

@article{hoope25ptchar_fromEHRs,
  author    = {ten Hoope, Simone and Welvaars, Koen and van Geijtenbeek, Kylian and Klok-Everaars, Mellanie and van Schaik, Sander and Karapinar-{\c{C}}arkit, Fatma},
  title     = {Applying text-mining to clinical notes: the identification of patient characteristics from electronic health records ({EHR}s)},
  journal   = {BMC Medical Informatics and Decision Making},
  volume    = {25},
  pages     = {302},
  year      = {2025},
  doi       = {10.1186/s12911-025-03137-x}
}

@article{weiss05_nvm_hlscreening,
  author    = {Weiss, Barry D. and Mays, Michael Z. and Martz, William and Castro, Karen M. and DeWalt, Darren A. and Pignone, Michael P. and Mockbee, John and Hale, Faye A.},
  title     = {Quick Assessment of Literacy in Primary Care: The Newest Vital Sign},
  journal   = {The Annals of Family Medicine},
  year      = {2005},
  volume    = {3},
  number    = {6},
  pages     = {514--522},
  month     = {Nov-Dec},
  doi       = {10.1370/afm.405},
  pmid      = {16338915},
  pmcid     = {PMC1466931}
}

@article{wallston14_briefhlscreen,
  author    = {Wallston, Kenneth A. and Cawthon, Curtis and McNaughton, Candace D. and Rothman, Russell L. and Osborn, Chandra Y. and Kripalani, Sunil},
  title     = {Psychometric Properties of the Brief Health Literacy Screen in Clinical Practice},
  journal   = {J Gen Intern Med},
  year      = {2014},
  volume    = {29},
  number    = {1},
  pages     = {119--126},
  month     = {January},
}

@article{mccray05promotinghealthlit,
    author = {McCray, Alexa T.},
    title = {Promoting Health Literacy},
    journal = {JAMIA},
    volume = {12},
    number = {2},
    pages = {152-163},
    year = {2005},
    month = {03},
}

@article{duell15optimal_hl_assessment,
title = {Optimal health literacy measurement for the clinical setting: A systematic review},
journal = {Patient Education and Counseling},
volume = {98},
number = {11},
pages = {1295-1307},
year = {2015},
author = {Paul Duell and David Wright and Andre M.N. Renzaho and Debi Bhattacharya},
}

@article{campbell19HL_structured,
  author    = {Campbell, Paul and Lewis, Martyn and Chen, Ying and Lacey, Rosie J. and Rowlands, Gillian and Protheroe, Joanne},
  title     = {Can patients with low health literacy be identified from routine primary care health records? A cross-sectional and prospective analysis},
  journal   = {BMC Family Practice},
  year      = {2019},
  volume    = {20},
  number    = {1},
  pages     = {101},
  issn      = {1471-2296},
}

@online{qwen3_32b_huggingface,
  author  = {{Tencent AI Lab}},
  title   = {Qwen3-32B},
  year    = {2024},
  url     = {https://huggingface.co/Qwen/Qwen3-32B},
  urldate = {2026-03-08},
}

@online{llama_8B_huggingface,
  author  = {{Meta AI}},
  title   = {Meta-Llama-3-8B-Instruct},
  year    = {2024},
  url     = {https://huggingface.co/meta-llama/Meta-Llama-3-8B-Instruct},
  urldate = {2026-03-08},
}

@online{llama70B_huggingface,
  author  = {{Meta AI}},
  title   = {Meta-Llama-3-70B-Instruct},
  year    = {2024},
  url     = {https://huggingface.co/meta-llama/Meta-Llama-3-70B-Instruct},
  urldate = {2026-03-08},
}

@online{qwen3_8b_huggingface,
  author  = {{Tencent AI Lab}},
  title   = {Qwen3-8B},
  year    = {2024},
  url     = {https://huggingface.co/Qwen/Qwen3-8B},
  urldate = {2026-03-08},
}

@article{chesser16healthlit_age,
  author  = {Chesser, Amy K. and Keene Woods, Nicole and Smothers, Kourtney and Rogers, Nancy},
  title   = {Health Literacy and Older Adults: A Systematic Review},
  journal = {Gerontology and Geriatric Medicine},
  year    = {2016},
  volume  = {2},
  pages   = {2333721416630492},
  doi     = {10.1177/2333721416630492}
}

\end{document}